# Autonomous Business System via Neuro-symbolic AI

Cecil Pang[1, 2]
[1]*School of Systems Science and Industrial Engineering, Binghamton University, State University of New York, Binghamton, USA*
[2]*AI Engineering, USA TODAY Co., Inc., New York, USA*
ORCID: 0009-0003-1725-7873

Hiroki Sayama[1, 3, 4]
[1]*School of Systems Science and Industrial Engineering, Binghamton University, State University of New York, Binghamton, USA*
[3]*Binghamton Center of Complex Systems, Binghamton University, State University of New York, Binghamton, USA*
[4]*Waseda Innovation Lab, Waseda University, Tokyo, Japan*
ORCID: 0000-0002-2670-5864

***Abstract*— Modern business environments demand continuous reconfiguration of cross-functional processes, yet most enterprise systems remain organized around siloed departments, rigid workflows, and hard-coded automation. Meanwhile, large language models (LLMs) demonstrate strong capabilities in interpreting natural language and synthesizing unstructured information, but they lack deterministic, auditable execution of complex business logic. We introduce Autonomous Business System (AUTOBUS), a system that integrates LLM-based AI agents, predicate-logic programming, and business-semantics-centric enterprise data into a unified neuro-symbolic architecture for executing end-to-end business initiatives. AUTOBUS models a business initiative as a network of interrelated tasks with explicit pre- and post-conditions, required data, evaluation rules, and API-level actions. Enterprise data is organized as a knowledge graph, whose entities, relationships, and constraints are translated into logic facts and foundational rules that ground reasoning and ensure semantic consistency. Core AI agents synthesize task instructions, enterprise semantics, and available tools into task-specific logic programs, which are executed by a logic engine that enforces constraints, coordinates auxiliary tools, and produces deterministic outcomes. Humans specify task instructions, define and maintain business semantics and policies, curate tools, and supervise high-impact or ambiguous decisions, ensuring accountability and adaptability. We detail the AUTOBUS architecture, the structure of AI-generated logic programs, and the human-AI collaboration model and present a case study that demonstrates accelerated time to market in a data-rich organization. A reference implementation of the case study is available at https://github.com/cecilpang/autobus-paper.**

## I. Introduction

### A. Background

In today's markets, businesses must continuously identify opportunities, adjust strategies, and optimize tactics at short notice in order to grow or simply remain competitive. Meanwhile business capabilities increasingly rely on the integration of diverse functions—sales, marketing, finance, logistics, and data—working together toward shared outcomes. Traditional enterprise software, however, is organized around siloed departments and optimizes tasks rather than end-to-end value creation. Even though Enterprise Resource Planning (ERP) software suites unify data across modular components [1], which helps with cross-module analysis and reporting, it often requires rigid workflows that impede rapid changes. As a result, business organizations struggle to reconfigure capabilities at the pace required by today's markets. This challenge sits at the core of enterprise digital transformation, which is more than technology adoption: it is a multi-dimensional shift spanning strategy, people, organizational structure, customers, ecosystem partnerships, technology, and innovation. Recent case-based research shows that successful digital transformations depends on coordinated, cross-functional orchestration rather than isolated tools [2].

Decades of business process redesign (BPR) and business process management (BPM) research anticipated this need. Classic BPR argues that companies should "reengineer" work—rethinking processes across functional boundaries instead of automating legacy tasks—because dramatic performance gains come from process-centric, not department-centric, design. Complementary work on process innovation positions technology as the key enabler of such redesign. Together, these perspectives emphasize end-to-end process orientation, explicit ownership of outcomes, and measurement against customer-facing results [3], [4]. More recent research links BPM and BPR directly to digital transformation as its integral part [5], [6]. It requires new BPM logics—lighter-weight processes, infrastructural flexibility, and "mindful actors" who can act on less than ideal situations—that better accommodate continual change and data-driven decision-making [7].

Recent advances in large language model (LLM)-based AI agents offer new opportunities to re-imagine how business operates. Multi-agent systems (MAS), once largely confined to scientific and engineering domains, are becoming viable in business contexts thanks to the increasingly software-centric and data-driven nature of business operations. A recent study by Dennis et al. [8] suggested that AI agents are likely to be accepted as full team members, requiring many classic collaboration questions to be re-examined in the context of human-AI teams. Saghafian and Idan [9] argued that the future application of AI in many domains should focus on combining artificial and human intelligence.

While LLMs excel in natural language understanding, learning and semantic reasoning over unstructured data, they often struggle with deterministic execution, especially when it involves complex and lengthy logic reasoning [10]. This makes them less effective in business process automation where decisions need to be reasonable and verifiable. This gap can be filled by a traditional branch of AI—predicate calculus-based logic programming [11]. The formal reasoning nature of logic programming provides a simple and compact mechanism to encode business logic, and it is intuitive to reason over. The combining of the strengths of neural network-based LLM and the reasoning power of logic programming, which is a type of symbolic AI, is known as a neuro-symbolic approach.

### B. *Proposed System*

We introduce Autonomous Business System (AUTOBUS)—a system that intelligently executes end-to-end business initiatives by integrating human intelligence, neuro-symbolic AI, and business semantics grounded in enterprise data. A business initiative represents a concrete objective with defined goals, evaluation metrics, constraints, and expected artifacts. In contrast to software systems organized around fixed and siloed departments, AUTOBUS treats the cross-functional human-AI teams as the primary building block of the enterprise, aiming to deliver end-to-end enterprise level results. *Figure 1* is an overview of AUTOBUS.

AUTOBUS is best suited for modern data-rich organizations that rely on cross-functional workflows. In such an environment, business initiatives are typically data and software driven, and operational systems expose APIs and events so that data can flow across tasks with limited manual intervention.

Existing neuro-symbolic approaches in literature that utilize LLMs and logic programming tend to focus on narrow reasoning tasks or domains. To the best of our knowledge, no prior work was designed for real-world, data-rich business environments, providing actionable guidance for the design and operation of end-to-end business initiatives. AUTOBUS:

- improves time to market: a coherent data, humans, and neuro-symbolic AI architecture enables business initiatives to iterate rapidly, and
- provides practical guidance**:** designed for real-world businesses, we provide reference implementation of a case study with open source code in GitHub. https://github.com/cecilpang/autobus-paper.

### C. *Related Work*

Research on autonomous business and agent-based business process management can be dated back to more than two decades. Jennings et al. [12] described an agent-based business process management system where agents are humans. Flor [13] explored programmable autonomous business built entirely in software and did not require humans to operate other than the community of customers of the business. More recently, Hughes et al. [14] investigated how AI agents and agentic systems could reshape industries by decentralizing decision-making, reconfiguring organizational structures, and strengthening cross-functional collaboration. Lashkevich et al. [15] studied using LLMs to assist the redesign of business process to reduce waiting time. Regarding having AI agents as team members, Dennis, Lakhiwal and Sachdeva [8] studied the potentially positive and negative disruptive influence on teamwork. Lemon [16] surveyed the challenges and some of the open research directions in multi-party systems where AI agents interact with more than one human.

Even though LLM-based agentic systems are a very recent development, they have already made tremendous progress. Acharya, Kuppan and Divya [17] surveyed the current and potential applications in various real-world domains such as healthcare finance, supply chain, education, and adaptive software systems. Khamis [18] examined the different architectural components of an agentic system. Hu, Lu and Clune [19] even proposed a new research area of automated design of agentic systems.

While neuro-symbolic AI has been gaining ground in AI research according to these review papers: [20], [21], [22], research on integrating LLM and logic programming for business systems has only recently emerged. Mezini et al. [23] integrated temporal-logic rules with learned neural network models to predict business process outcomes. Manuj et al. [24] and used LLMs to extract facts from legal contracts and then used logic programs for coverage and eligibility reasoning for policy compliance and claims. Neuro-symbolic AI has also been applied in E-commerce recommendation systems [25], [26], [27]. Though not specifically for business systems, Vakharia et al. [28] used logic programming to improve the performance of LLMs in domain specific question answering tasks. Pan et al. [29] used LLMs to translate logical reasoning problems in natural language into a symbolic representations and then used a symbolic logic solver to solve the problems deterministically. Borazjanizadeh and Piantadosi [30] showed that integrating logic programming into the inference pipeline of LLMs yielded robust performance gains. Yang, Chen and Tam [31] used logic programs generated by LLMs to solve arithmetic problems.

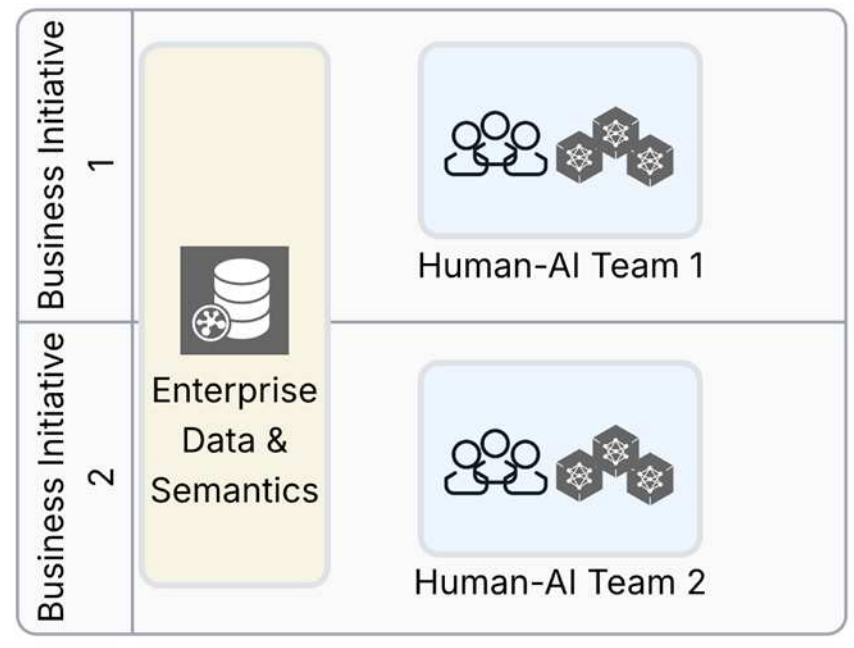

Fig.1. Overview of Autonomous Business System (AUTOBUS). AUTOBUS integrates human intelligence, neuro-symbolic AI, and business semantics to orchestrate end-to-end business initiatives. Teams in which humans and AI are peer members collaborating under shared objectives, metrics, constraints, and expected artifacts—delivering enterprise-level outcomes.

## II. AUTOBUS ARCHITECTURE

### A. *Business Initiative*

A business initiative is modeled as a sequence of tasks, often with parallel branches, that transform inputs into measurable outcomes, as illustrated in *Figure 2*. Each task has explicit pre-/post-conditions, required data, and well-defined actions. Typically these actions are executed by operational software such as API calls. Most importantly, decisions were made throughout the execution of the tasks. Each individual task as well as the business initiative as a whole are evaluated by defined rules and metrics. Parts or the whole of the initiative can be iterative.

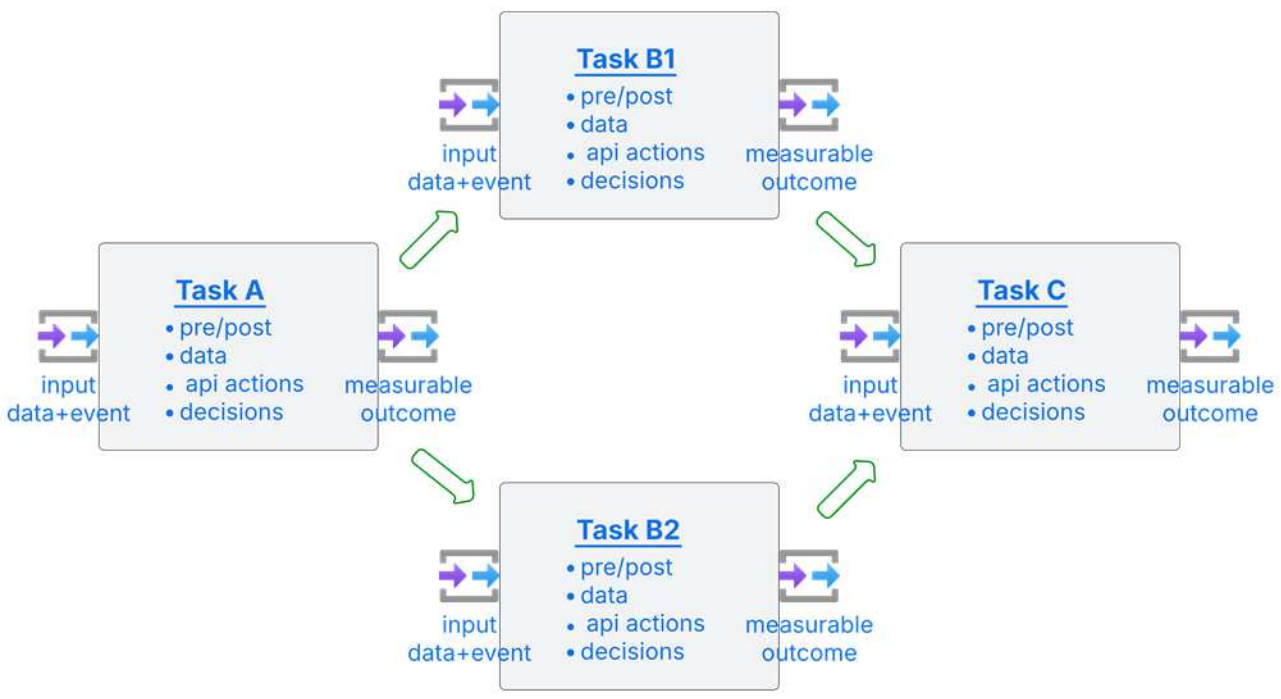

Fig. 2. Business initiative model: inputs traverse a sequence of data-driven tasks—often with parallel branches—to yield measurable outcomes. Each task defines pre/post-conditions, required data, and software-executed API actions, with decisions throughout. The initiative is evaluated by goals and metrics, and sub-flows may iterate.

## B. *Principal Components*

At the core of AUTOBUS are humans, enterprise data, LLM-based AI agents, and a logic engine. Each business initiative has its own task- and domain-specific instructions, while auxiliary tools—including other AI agents, machine learning models, and internal or external APIs—provide additional computational and informational support, as well as sending actions to operational systems. *Figure 3* presents the principal components of AUTOBUS.

Operationally, the core AI agents construct logic programs by synthesizing information from business semantics, task instructions and tools availability. These logic programs are then executed by the logic engine. Groundings for facts are typically from enterprise data. In addition, the logic programs invoke tools to get groundings for predicates that cannot be obtained upfront from enterprise data. Some examples are real-time data from the web, data science models, and system events. The logic programs could also invoke other AI agents that in turn utilize other tools or resources to fulfill the request. Tools are also used to perform actions of the logic programs, such as calling APIs or persisting outcome of the logic program. *Figure 4* depicts the following steps:

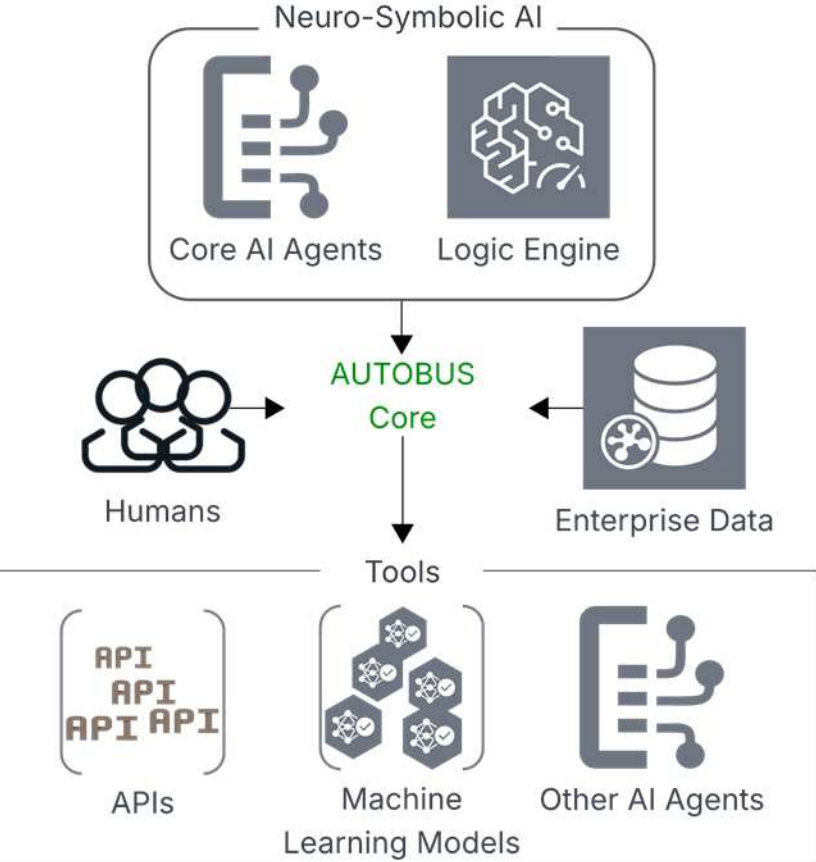

Fig.3. Principal components of AUTOBUS. Neuro-symbolic AI, humans, and enterprise data provide complementary intelligence and context to the AUTOBU core, which coordinates decisions and actions. A tooling layer of APIs, AI agents, machine-learning models, and other integrations executes these decisions against operational systems.

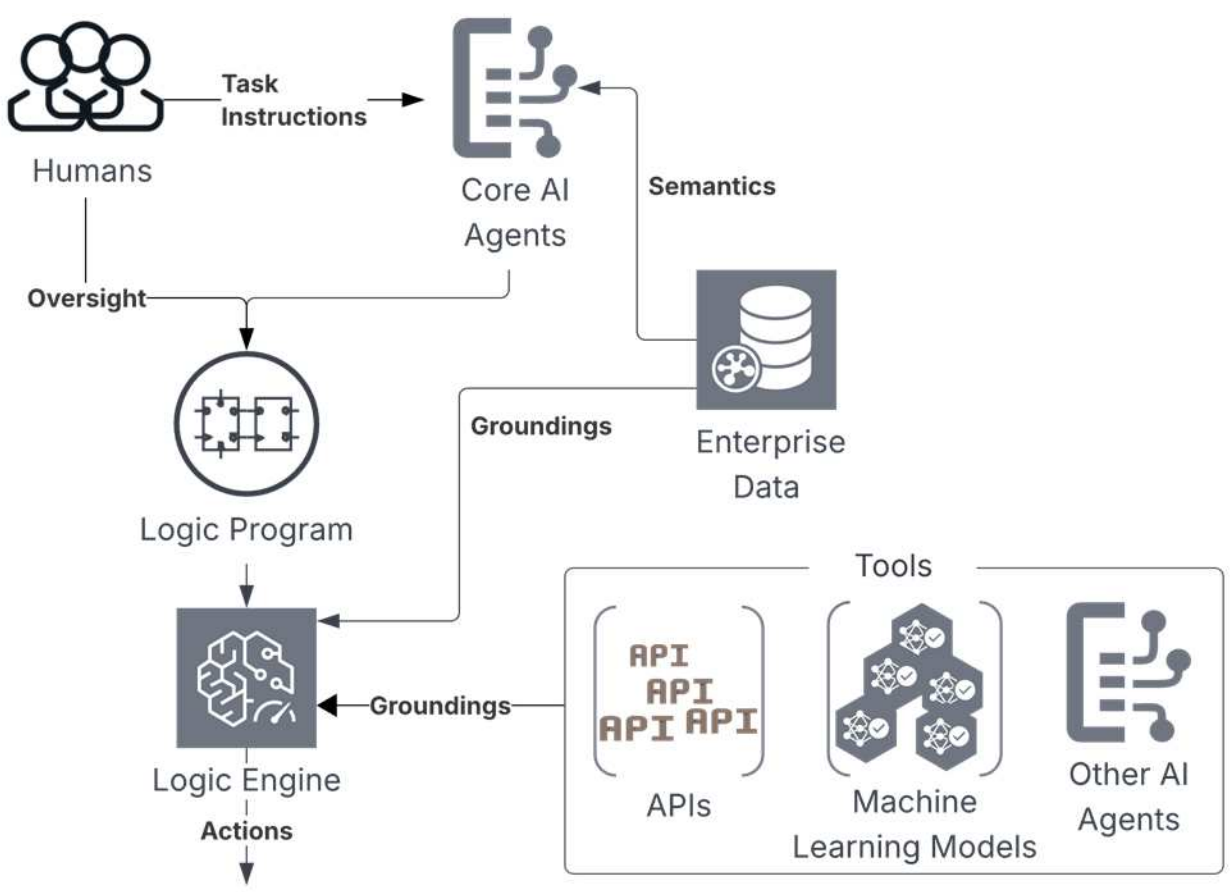

Fig. 4. Logic program generation and execution. AI agents translate task instructions into logic predicates and enrich them with relevant logic facts, foundational rules, and groundings derived from enterprise data and auxiliary tools. The resulting logic program is executed by the logic engine, which determines task feasibility, enforces constraints, and orchestrates actions.

- translate task instructions into logic predicates,
- augment these predicates with the relevant facts and foundational rules derived from business semantics in enterprise data,
- specify groundings from enterprise data and auxiliary tools such as machine learning models, API calls and other AI agents, and
- humans review the generated logic and provide oversight for high-stakes decisions.

The logic engine then executes the generated logic program to drive actions.

## C. *Logic Program*

Logic programs model tasks using the fundamental constructs of predicate logic—the language of facts, rules, and structured inference. SWI-Prolog [32] is used in this work. As its author stated, SWI-Prolog is an integration tool, which is an important role of logic program in AUTOBUS. At a conceptual level, a task can be described by logic predicates such as:

```
task(T).
requires(T, Data).
precondition(T, P).
postcondition(T, Q).
```

These predicates capture the essential semantics of the tasks, their dependencies, and their expected effects. They allow AUTOBUS to represent a business initiative not as ad-hoc procedural code but as a declarative structure that explicitly states what must be true before and after each task, as well as what data and decisions govern its execution. This makes the intent and logic of the tasks transparent, auditable, and directly aligned with business semantics.

Building on these preliminaries, logic programming enables AUTOBUS to orchestrate tasks by reasoning over the evolving state. For example, a task is executable when the logic inference engine can derive that all its preconditions hold. Formally, when predicates such as

```
holds(P, State).
```

are satisfied, the rule that governs task readiness is satisfied. Parallel branches arise naturally as multiple tasks simultaneously satisfy their preconditions, and iterative structures can be expressed through rules that re-activate a task when certain conditions recur:

```
repeat_until(Task, Goal).
```

Decisions within a task are represented by rules with alternative heads or guarded conditions, allowing the system to select actions, such as invoking an API to update enterprise data based on the current logical state:

```
invoke(Action, Params).
```

Evaluation rules and metrics are encoded as higher-level predicates that compute whether an initiative or a task meets defined business criteria, enabling continuous assessment during execution. For example, a customer success evaluation of whether a business process resolves the customer problem combines operational and customer-satisfaction metrics into a single evaluation rule:

```
resolved(I) :- outcome(I, resolved).

success(I) :-
    resolved(I),
    customer_satisfaction(I, Score),
    Score >= 4.0.
```

A logic program for a task in AUTOBUS consists of three main parts: 1) facts and foundational rules, 2) task specific rules, and 3) predicates for actions and task outcomes. Enterprise data serve as the source of logic facts and foundational rules, which describe business entities, their relationships, and constraints. These are common to all initiatives within an organization and serve as the anchor, providing a shared foundation for tasks within an initiative, and across all initiatives. To enable AI agents to reliably derive these facts, AUTOBUS adopts the business-semantics–centric enterprise data system specified in a recent work [33]. Guided by task instructions, the AI agents extract and construct the facts and foundational rules, grounding them in the enterprise's formal semantic structures. This semantic foundation is essential for ensuring that the generated logic programs are consistent with domain meaning and for maximizing automated logic program generation and minimizing the need for humans to describe the data manually. These enterprise-data-derived facts can be further augmented with real-time information, including human inputs, live machine learning model inferences, API responses, and outputs produced by other AI agents.

Beyond the facts, there are task-specific rules. These rules are generated by the AI agents based on the task instructions and descriptions of various tools available. Since tasks differ in purpose, scope, and domain, the resulting rules can vary widely. Common categories include rules that identify and filter business entities based on task-specific criteria, and provide reasoning from facts via rules to actions.

For predicates that lead to actions or task outcomes, it involves triggering some events. For modern data- and software-driven businesses, it typically means invoking the services of some operational systems, passing in some prepared dataset. It could also simply be persisting the outcomes of the task. *Figure 5* illustrates the anatomy of a logic program in AUTOBUS.

### D. *Business Semantics Centric Enterprise Data*

A foundational component of AUTOBUS is a business semantics centric enterprise data system, as specified by Pang [33]. Such a system is an essential and integral element of a modern data-driven business rather than merely a data management utility. We extend this view by positioning this enterprise data system as one of the three pillars of autonomous business, together with humans and neuro-symbolic AI. The enterprise data system organizes information into typed and schema-governed structures around business entities such as consumers and subscriptions, and their relationships. These structured data assets form the basis of an enterprise knowledge graph (KG) in which entities, relationships, and constraints are explicitly represented. For example, consumer–subscription associations and status transitions are captured as KG triples aligned with the business organization's ontology and reference data models. This provides the business semantic from which AUTOBUS derives the facts and foundational rules, as well as their groundings, for the logic programs. KG triples are translated to logic facts, such as the following snippet, which states that `c123` is a consumer who subscribes to an active subscription `s456`:

```
consumer(c123).
subscription(s456).
subscribe(c123, s456).
has_status(s456, active).
```

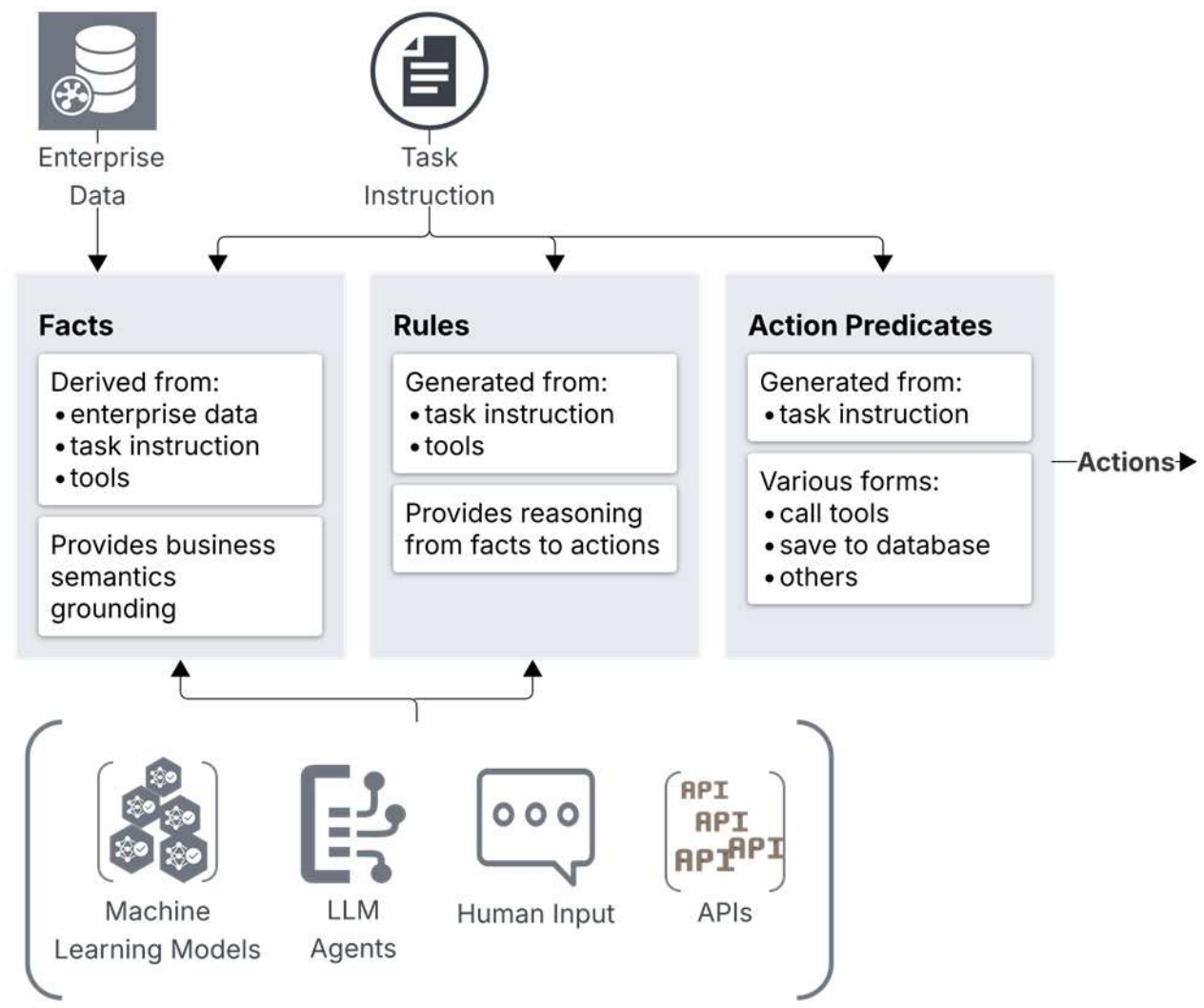

Fig. 5. Anatomy of an AUTOBUS logic program. Each task-level logic program comprises three elements: (1) facts and foundational rules derived from enterprise data, (2) task-specific rules generated from the task instructions and available tools, and (3) predicates that define actions and task outcomes. Task-specific rules identify relevant entities, apply filters, and connect facts to actions, while action predicates trigger operational events.

Schema-level constraints and domain rules in the KG are similarly translated to foundational logic rules, including integrity checks and task preconditions:

```
active_subscription(S):-
    has_status(S, active),
    subscribe(_, S).

precondition(send_promotion(C)):-
    consumer(C),
    subscribe(C, S),
    active_subscription(S).
```

The first predicate defines an active subscription as one that is subscribed by any consumer and has an active status. The second one states that promotion is sent to a consumer only if the consumer has an active subscription.

Structured enterprise data represent a semantic knowledge graph. When translated to logic facts and foundational rules, it provides the grounding for reasoning about the business. Methodologies for generating knowledge graph from relational data are well-established [34]. Typically, table rows are translated to graph nodes, which are then linked to their column values to form bipartite "star" sub-graphs. These sub-graphs are subsequently connected via shared values to form the entire graph.

### E. Humans

Humans play a key role in AUTOBUS, providing the domain expertise, contextual judgment, and strategic oversight that anchor the system's neuro-symbolic reasoning. Humans supply the nuanced judgements and interpretations—such as ethical considerations, policy intent, organizational priorities, and exception handling—that neither logic programs nor AI agents can reliably infer in isolation. Before setting up tasks for a business initiative, humans define and maintain the business semantics and structure enterprise data accordingly, and curate the ontologies and common policies that govern the logic of tasks. When setting up the business initiatives, humans compose task instructions, set the objectives and evaluation criteria, and curate tools to be used for the tasks. During execution of tasks, humans validate ambiguous or high-impact decisions in the generated logic programs, and iteratively refine the task instructions in response to task and initiative level outcomes. Human involvement is critical not only for ensuring correctness, accountability, and alignment with organizational goals, but also for enabling AUTOBUS to adapt to real-world complexity and maintain trustworthiness over time.

### F. Scalability Challenges and Strategies

Enterprise data in modern organizations can be massive, making it impractical to naively "dump" all entities, attributes, and relationships into a fully materialized set of predicate-logic facts for each initiative. Scalability challenges arise both in semantic grounding (extracting the *task-relevant* slice of the enterprise knowledge graph) and in logic execution (evaluating rules over large, evolving state). A practical strategy is therefore to adopt task-scoped, on-demand grounding: retrieve and materialize only the entities and relations required by a given task, leverage indices and schema/ontology constraints to prune the search space and incrementally update facts as new events arrive rather than rebuilding the fact base. On the execution side, logic programs can be compiled or pushed down to scalable backends—e.g., graph-native reasoning engines such as PyReason [35] that operate directly over graph databases, or declarative systems such as Logica [36] that compile to SQL for execution on distributed data platforms—enabling parallel evaluation, incremental maintenance, and cost-based optimization. While a full scalability evaluation is outside the scope of this paper, these strategies indicate a path to deploying AUTOBUS over large enterprise datasets and complex, real-world process graphs.

### G. Trust and Governance

AUTOBUS is designed to make long-horizon business automation trustworthy, governable, and auditable under realistic failure conditions. Expected failure modes include (i) LLM synthesis errors (e.g., missing constraints, incorrect tool selection, or inconsistent predicates), (ii) incomplete/stale enterprise facts, and (iii) downstream tool/API/model failures. AUTOBUS mitigates these via an explicit logic layer with constraint checks, provenance tracking, and policy gates, and by routing low-confidence or high-impact actions to human review that can lead to a "break-glass" override with full justification capture.

This trust and governance emphasis also highlights where prior approaches fall short for practical business initiatives: pure LLM tool-use agents can be highly automated but often lack deterministic behavior, stable intermediate artifacts, and end-to-end audit trails; classical BPM/RPA pipelines provide operational control and logging but tend to embed decision logic in brittle scripts that are hard to validate semantically across branching/iterative workflows; and pure rule engines are verifiable but require costly manual knowledge engineering and struggle with unstructured inputs and tool-rich execution. By grounding tasks in enterprise semantics (for privacy-preserving, least-privilege fact access) and representing decisions as a versioned, testable logic program with traceable rule firings and tool calls, AUTOBUS advances beyond prior work in the non-functional properties that are essential for real-world business deployments: auditability, accountability, change management, and compliance-ready verification.

## III. Validation Case Study

### A. Introduction

We validate the effectiveness of AUTOBUS through a case study of implementing a proof of concept for an online content subscription business. We break down the business initiative into three tasks, and describe the generated logic programs for them. We then compare and contrast with the traditional approach and provide metric that shows that AUTOBUS accelerates the time to market. We also discuss the factors contributing to this improvement.

As presented in *Figure 6*, the key business entities involved in the case study are consumer, subscription, and product. A consumer is either a subscriber, who subscribes to one or more products, or a non-subscriber. There are profile attributes attached to consumers, including information about the consumers collected from the subscriptions as well as data from third-party data providers.

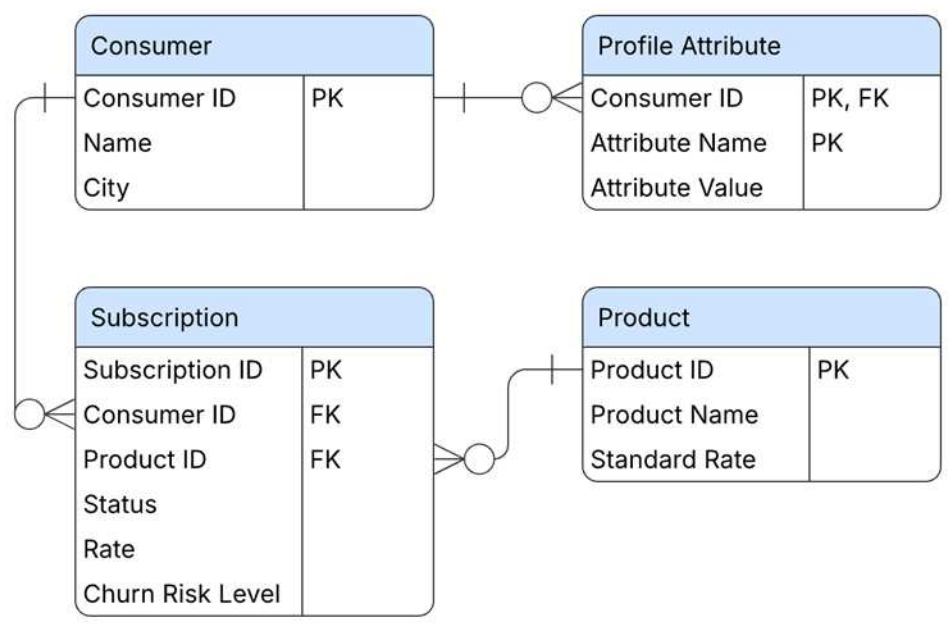

Fig. 6. Entity–relationship diagram for the case study. The primary business entities are consumer**,** subscription**,** and product. Consumer profile attributes obtained from subscription data and augmented with third-party data sources.

## B. *Business Initiative and Tasks*

For a subscription-based business, the top priority is to acquire new subscribers and at the same time retain existing ones. This case study is about subscriber retention. A common tactic is to proactively reach out to high-churn-risk subscribers and offer them perks. The perks can be for certain markets and/or specific products. Concretely, the case study is to send promotions to active subscribers who:

1. subscribe to a certain product, pay a monthly rate above certain dollars, have a certain churn risk level, and
2. have household incomes above the median of the cities residing in.

The initiative is broken down into three tasks, as in *Figure 7*:

1. Identify and retrieve the target subscribers based on Criterion 1 only.
2. Obtain the median household incomes for the cities where the subscribers reside. This involves calling a specialized AI agent that fetches median city household income data from the web.
3. Filter the subscriptions to include only those whose households have income above the cities' median. Then send them to the marketing platform.

## C. *Logic Programs*

The logic programs of the three tasks are generated by the same core AI agent of AUTOBUS, each with task specific instructions. Each logic program comprises three sections: (1)

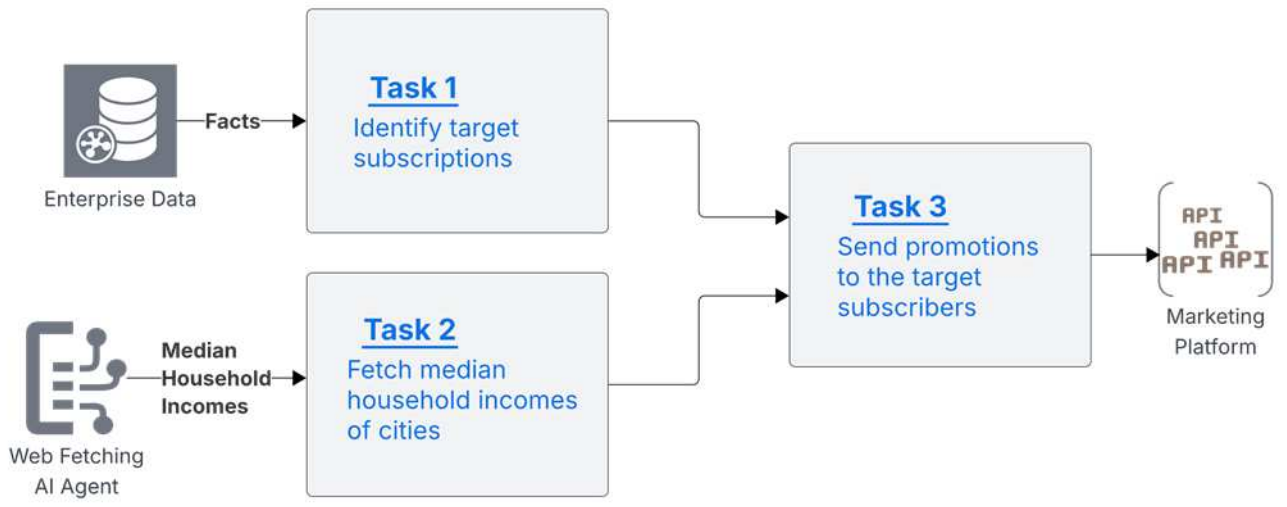

Fig. 7. The initiative targets subscribers who (i) subscribe to Product 1, (ii) have churn-risk level 4, and (iii) have household income above median of the city. It is organized into three tasks: retrieving eligible subscribers from enterprise data, fetching median incomes of cities from the web, and sending promotions to the resulting target subscribers through marketing APIs.

facts, (2) task specific rules, and (3) actions. These are typical elements of data-driven business tasks. The structure of the logic program generated for Task 3 is presented in *Figure 8*. It contains the following three elements:

1. Facts and foundational rules: loaded from enterprise data, and the outcomes of Task 1 and 2.
2. Task rules: the logic of targeting subscribers who (i) are savable churns determined by Task 1, and (ii) have household incomes above the medians of the cities of residence, which are the outcome of Task 2.
3. Actions: (i) persist the target subscriptions into the database, and (ii) send the target subscriptions to a marketing campaign via a tool call.

## D. *Result and Discussion*

The case study was conducted by the first author at the workplace as a proof of concept to mimic a typical business initiative that involves customer segmentation and engagement. While the ultimate goal is to improve retention, a useful metric is time-to-market. Being able to quickly iterate the initiative with controlled variations of parameters so as to validate hypothesis is a critical factor of minimizing churn. This metric is closely aligned with the main value proposition of AUTOBUS.

To evaluate time-to-market, we compare the time required to complete the three tasks shown in *Figure 7* with a baseline reflecting the organization's existing approach. In the baseline, each task requires building task-specific data pipelines and deploying them into the enterprise data infrastructure. The estimate is derived from observations of similar efforts conducted in the past, which typically involve multiple teams. Coordination across these teams typically requires several rounds of discussions before roles, responsibilities, and technical details are fully aligned. This overhead is repeated for subsequent initiatives—even when they are similar—because team composition often changes and time is again spent re-establishing shared understanding, down to minor details. The following is what contributes to the effort of each of the tasks of the baseline: 1) Retention team make a request to enterprise data management (EDM); 2) EDM leaders review intake requests and assign tasks to teams and engineers within EDM; 3) Data engineers meet with retention team to clarify requirements; 4) Engineers code data pipeline and deploy them to production infrastructure.

With AUTOBUS, task logic programs are generated by a consistent set of core AI agents using task instructions, shared enterprise data, and standardized tools. This approach largely eliminates the unproductive coordination cycles inherent in the baseline workflow. The reduction in redundant planning and alignment effort, and the consistent execution of the logics, are the primary contributors to the improved time to market.

The results are recorded in *Table 1*. Since Tasks 1 and 2 can be executed in parallel, the time to market is two days for AUTOBUS versus 2 weeks with the existing approach. There is flexibility in structuring the initiative. Breaking it down into three tasks in this case study is a design choice that follows a good system design practice of grouping logically related activities in separate modules for ease of reasoning, maintenance, and reuse. Functionally, it would work just as well if the three tasks are combined into one.

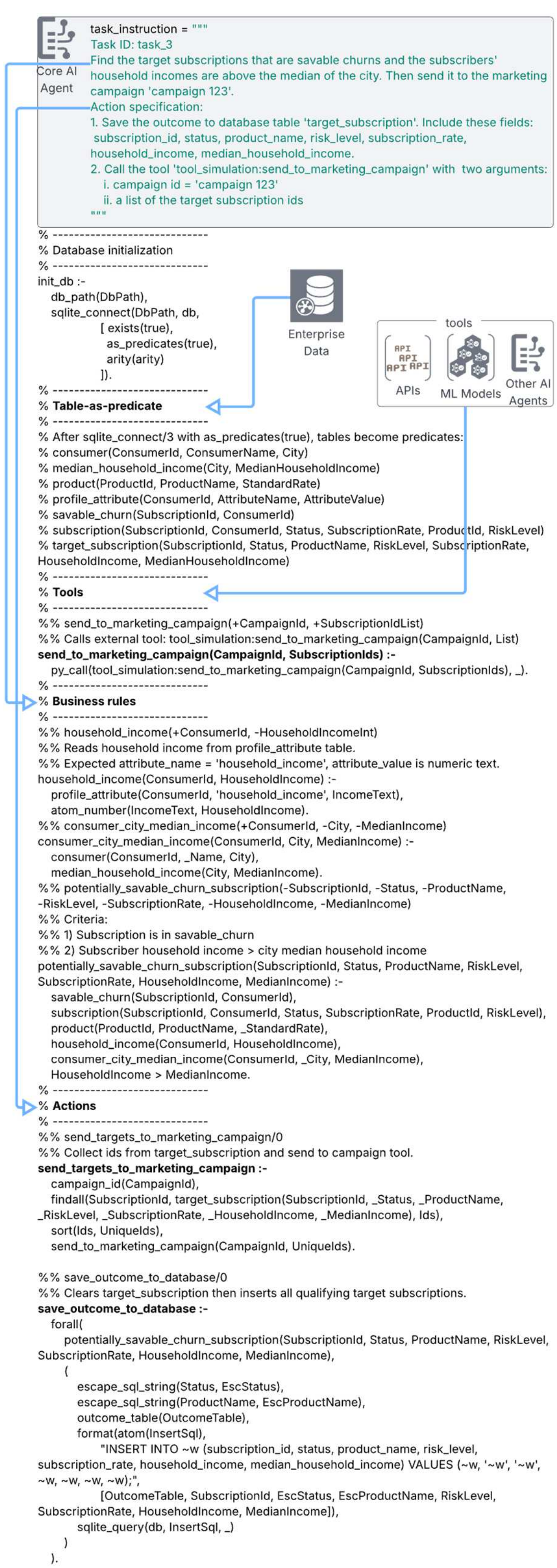

Fig. 8. Structure of the logic program for Task 3, generated by the AUTOBUS core AI agent from task-specific instruction. The program consists of three sections—facts, task-specific rules, and actions—where facts are derived from enterprise data and the outputs of Tasks 1 and 2; task rules encode the targeting logic; and actions persist the results and invoke marketing campaign APIs.

In principle, the AI agent could bypass the logic layer and directly generate SQL from enterprise data. While sufficient for simple tasks, this approach becomes difficult to validate and audit for complex, multi-step initiatives. The logic program provides a transparent and modular intermediate representation that supports constraint enforcement and systematic reasoning, as well as auditing and governance.

TABLE 1
CASE STUDY METRIC: TIME TO MARKET

| Task | | Existing Approach | AUTOBUS |
|---|---|---|---|
| 1 [a] | Retrieve savable churns based on product, rate and risk level | 1 week | 1 day |
| 2 [a] | Obtain the median household income for cities from the web | 1 week | 1 day |
| 3 | Further filter by household income and send result to marketing platform | 1 week | 1 day |

End-to-end case study execution time of AUTOBUS, with benchmark against the existing approach.

[a] Tasks 1 and 2 are run in parallel.

## IV. CONCLUSION, LIMITATION & FUTURE WORK

Autonomous Business System (AUTOBUS) provides a logical, clear, and well-organized architecture for businesses to focus their resources on end-to-end business initiatives that bring measurable values, rather than siloed departmental operations. It is also a practical system for humans to team up with neuro-symbolic AI in a data-rich business environment.

This work focuses on the architectural design and conceptual framework of AUTOBUS rather than a comprehensive empirical evaluation. As such, we do not report additional quantitative metrics such as logic execution time, error rates, rollback frequency, cost, or resource utilization. The evaluation is limited to a single case study centered on time-to-market acceleration and does not yet include additional case studies or synthetic benchmarks. Furthermore, broader validation dimensions—such as the accuracy of task-logic generation, reduction in human effort, system robustness under failure scenarios, or qualitative feedback from human participants—are not systematically assessed. We leave these empirical investigations to future work. We also do not address the organizational issue of transforming existing workforces to team with AI. This is a research topic on its own.

Lastly, we suggest an avenue for future work: priority-driven autonomous planning and optimization of business initiatives. Modeling initiative-level logic programs as graphs enables analysis and optimization using techniques from complex systems and adaptive network theory [37].